\documentclass[conference]{IEEEtran}
\IEEEoverridecommandlockouts
\usepackage{cite}
\usepackage{amsmath,amssymb,amsfonts}
\usepackage{algorithmic}
\usepackage{graphicx}
\usepackage{mathtools}
\usepackage{textcomp}
\usepackage{xcolor}
\usepackage{url}
\usepackage[affil-it]{authblk}
\def\BibTeX{{\rm B\kern-.05em{\sc i\kern-.025em b}\kern-.08em
    T\kern-.1667em\lower.7ex\hbox{E}\kern-.125emX}}
\begin{document}

\title{\textit{OysterSim}: Underwater Simulation for Enhancing Oyster Reef Monitoring\\
\thanks{Corresponding author: Xiaomin Lin. xlin01@umd.edu}
}

\author{Xiaomin Lin, Nitesh Jha, Mayank Joshi, Nare Karapetyan, Yiannis Aloimonos, and Miao Yu}
\affil{University of Maryland Institute for Advanced Computer Studies}

\maketitle

\begin{abstract}
Oysters are the living vacuum cleaners of the oceans. There is an exponential decline in the oyster population due to over-harvesting. With the current development of the automation and AI, robots are becoming an integral part of the environmental monitoring process that can be also utilized for oyster reef preservation. Nevertheless, the underwater environment poses many difficulties, both from the practical - dangerous and time consuming operations, and the technical perspectives - distorted perception and unreliable navigation.
To this end, we present a simulated environment that can be used to improve oyster reef monitoring. The simulated environment can be used to create photo-realistic image datasets with multiple sensor data and ground truth location of a remotely operated vehicle(ROV). Currently, there are no photo-realistic image datasets for oyster reef monitoring. Thus, we want to provide a new benchmark suite to the underwater community.
\end{abstract}


\section{Introduction}

Oysters are an essential species in the Bay living ecosystem. They are the living filters for the Chesapeake Bay. Oyster reefs have undergone major devastation due to over-harvesting. To accelerate their preservation, systematic monitoring of oyster reefs is essential. With the advancement of robotics and artificial intelligence technologies, we are able to use Remotely Operated Vehicles (ROV) or autonomous robots to monitor the aquatic environments\cite{manjanna2016efficient,manderson2017robotic,karapetyan2018multi,karapetyan2019riverine,karapetyan2021meander}. 

To fully automate the monitoring process though, localization capability is of vital importance which is especially challenging for the underwater domain\cite{quattrini2016experimental}. Moreover, efficient mapping and recognition approaches that are targeted specifically for oyster reefs will be a great help and can also assist the localization. These problems however are bounded by the existence of the oyster data, collection of which with underwater robots is a costly and labor-intensive task. 

To facilitate the development of underwater localization, recognition, and planning algorithms for oyster monitoring, we present a novel open-source simulation software\footnote{\url{https://github.com/prgumd/Oystersim.git}} and render datasets collected by the BlueROV robot underwater with multiple sensor data. This project is conducted in cooperation with Smart Sustainable Shellfish Aquaculture Management (S$^3$AM) which has a primary objective of enhancing oyster habitat.

\begin{figure}
 \centering
{\includegraphics[width=0.5\textwidth]{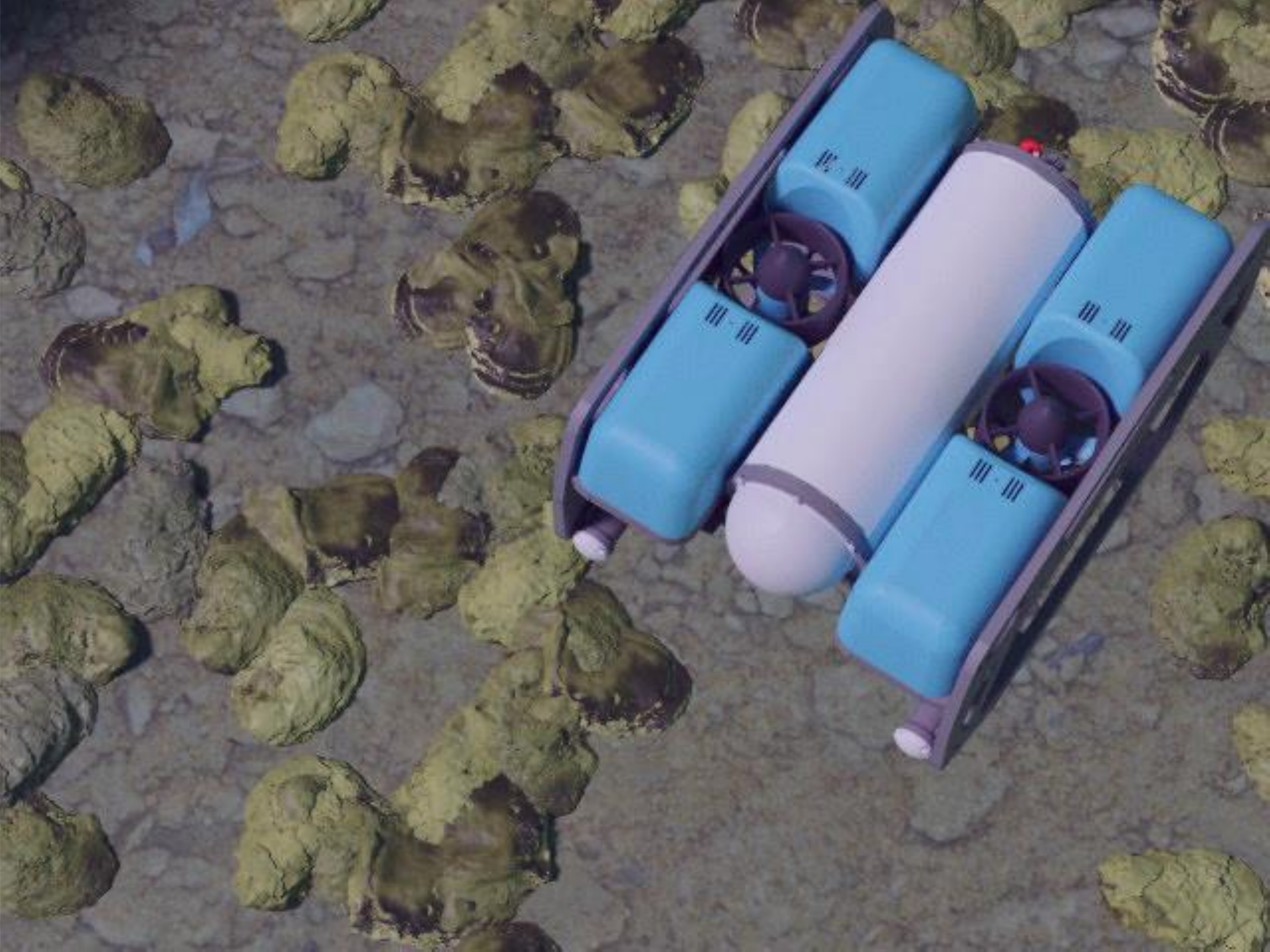}}
\caption{\label{fig:p1} Top-down view of a BlueROV navigating over an oyster reef in \textit{OysterSim} environment.}
\label{fig:oysteranddrone}
\end{figure}

 Oyster detection has been studied by Sadrfaridpour et al.\cite{sadrfaridpour2021detecting} opening up possibilities for automation of oyster reef monitoring. To have a robust system, extensive experiments must be performed with robotic systems in oyster environments. The objective of this work is to create a simulator that enables synthetic underwater dataset production for rover-based oyster monitoring which includes sensor data and ground truth. 
 
 There are a number of underwater simulations that provide advanced marine dynamics and different features for deploying a variety of  robots~\cite{perez2013underwater,henriksen2016uw}, nevertheless, they are not tailored toward specific environmental feature, which in our case is the oyster bed. To the best of our knowledge, there are no synthetic underwater datasets for rover-based oyster monitoring with close-to-reality images and ground truth data. Thus, we want to propose \textit{OysterSim} - a new open-source simulation to the underwater engineering community. We model the BlueROV as a point object because our task does not require aggressive maneuvers, and it is mainly in shallow water. We embedded real oysters at the bottom of the seabed of the simulation. 
 
 The rest of this paper is structured as follows. We will first give an overview of the literature relevant to this work in Section\ref{section:lit}. Then Section\ref{section:proposed} presents in details our proposed simulation framework - \textit{OysterSim} and it's components. In Section\ref{section:app} we demonstrate several applications that successfully utilized \textit{OysterSim}. Finally the concluding thoughts are drawn in the Section\ref{section:conclusion} with future work remarks. 
 
\section{Related Work}
\label{section:lit}
Simulation environments are an important part of any robotic application development. Even more, when operating in complex and dangerous environments the simulation is oftentimes the only tool available to testing robotic application. This is particularly true for the underwater applications. As such different research groups have develop underwater simulations to meet their needs.

Zwilgmeyer et al. \cite{zwilgmeyer2021creating} use Blender\cite{blender2018blender} to create underwater data containing ground truth and sensor data. A framework for the generation and measurement of sensor trajectories was created. To create paths, they use a physical model of the vehicle coupled with a control system that follows a predetermined path in a virtual underwater environment. Simultaneously, the state generated by the control system is used to create combined measurements of acceleration, angular velocity, and depth. The underwater scene was filled with a variety of geometries to create scenes with features ranging from simple sandy bottoms to vertical rocks to complex geometries with significant amounts of occlusion. Similar to the game engine used by Zwilgmeyer et al., our simulation is also based on Blender, but in contrast it is tailored towards oyster reef environments.

Some works, instead of utlizing Blender have successfully built their simulation packages with Gazebo platform\cite{koenig2004design}.  For example, UUV simulator\cite{manhaes2016uuv} is based on ROS and uses the Gazebo simulator. It models underwater physics precisely and also has a prototype SONAR implementation. But it requires ROS to be set up. The UUV Simulator SONAR model makes use of a simulated depth camera and GPU calculations are promising but has limitations because the depth camera's range of view does not correspond to that of a real imaging sonar.

Based on OpenSceneGraph\cite{wang2010openscenegraph} and osgOcean\footnote{\url{http://code.google.com/p/osgocean}}, the UWSim \cite{perez2013underwater} was created. It supports several agents, is open source, and has a sonar model that is more equivalent to a LiDAR, however, it is difficult to configure, dependent on ROS, and is not being actively maintained.
A currently well-maintained open-source robotics simulator based on reinforcement learning is called Holodeck \cite{HolodeckPCCL}. It is based on UE4 (Unreal Engine 4), giving it access to high-fidelity images, precise dynamics built on the PhysX physics engine, and an established community with many environment assets already created.
HoloOcean \cite{potokar2022holoocean} builds on Holodeck, with augmentations in underwater dynamics. It is open source, built upon UE4. HoloOcean supports multiple agents and has various sensor implementations of common underwater sensors and simulated communications support.
\begin{figure}[h!]
		\centering      
		\includegraphics[width=\linewidth]{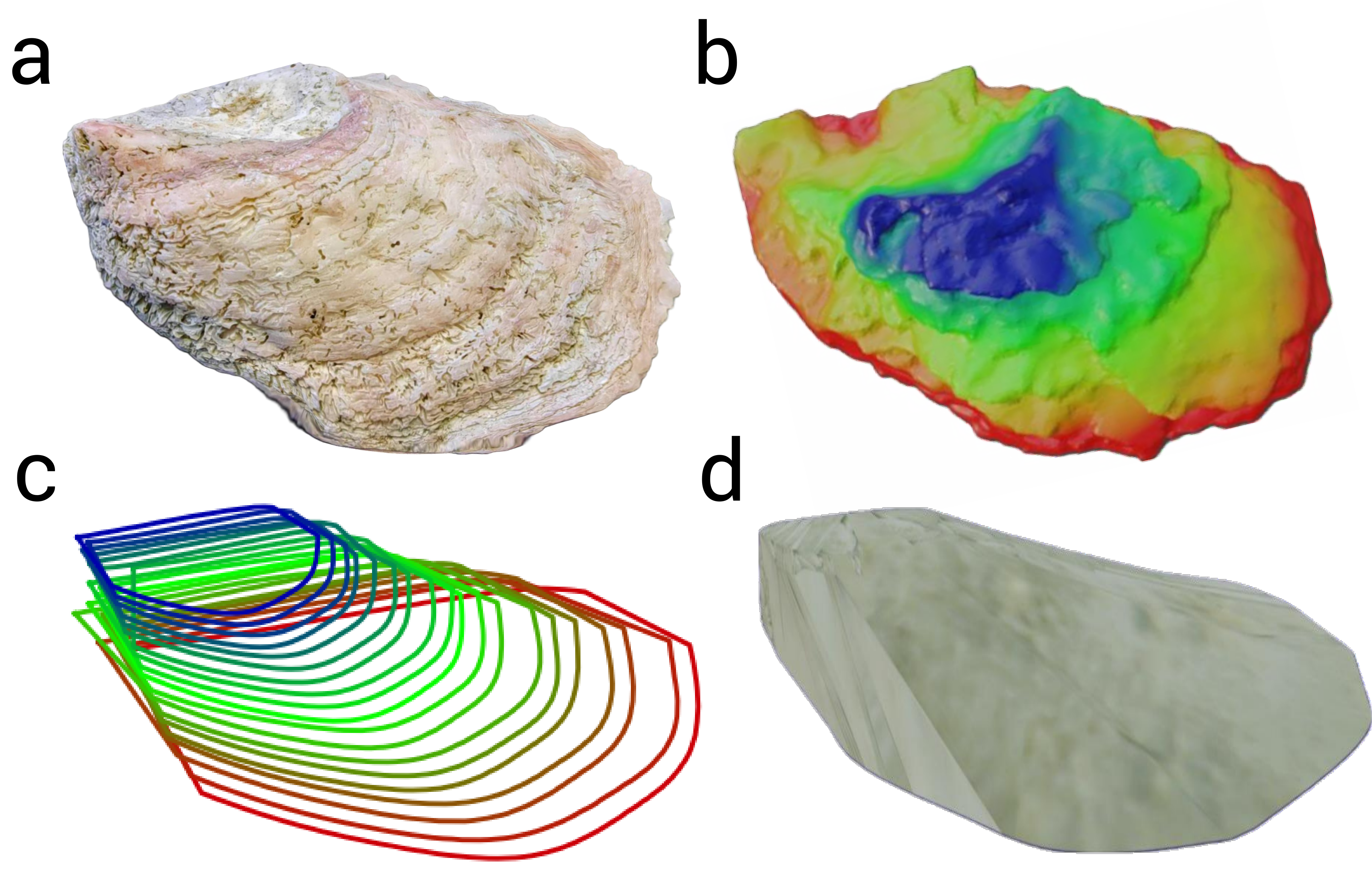}
		\caption{(a) Real oyster shell (washed). (b) 3D Scan of a real oyster shell. (c) Layered representation of a modeled oyster shell. (d) 3D model generated from the modeled oyster shell with a texture.}
		\label{fig:real-2-sim}
	\end{figure}

\section{PROPOSED APPROACH}
\label{section:proposed}
In this section, we first will present the  oyster models and the different sensors integrated into the \textit{OysterSim}. Following we will discuss the effect of water in \textit{OysterSim}. Last but not least, we present some details of the simulated scenes in Blender.

\subsection{Oyster modeling}
In order to model the geometry of an oyster, we first scanned ten washed oysters (Figure~\ref{fig:real-2-sim}(a)) to obtain their 3D-model (Figure~\ref{fig:real-2-sim}(b)). We modeled the oysters using a set of mathematical functions in a layered fashion (similar to the 3D printer's printing approach). Each layer - horizontal slice of the cross-section of the oyster, is modeled by two separate B-Spline as shown in Figure~\ref{fig:real-2-sim}(c). Then we extrude the 3D model from the layers by using the pyvista~\cite{sullivan2019pyvista} data visualizer. Following this step, we added the texture to the generated 3D model, which resulted in the osyter model depicted in the Figure~\ref{fig:real-2-sim}(d).

\begin{figure}[b!]
		\centering      
		\includegraphics[width=\linewidth]{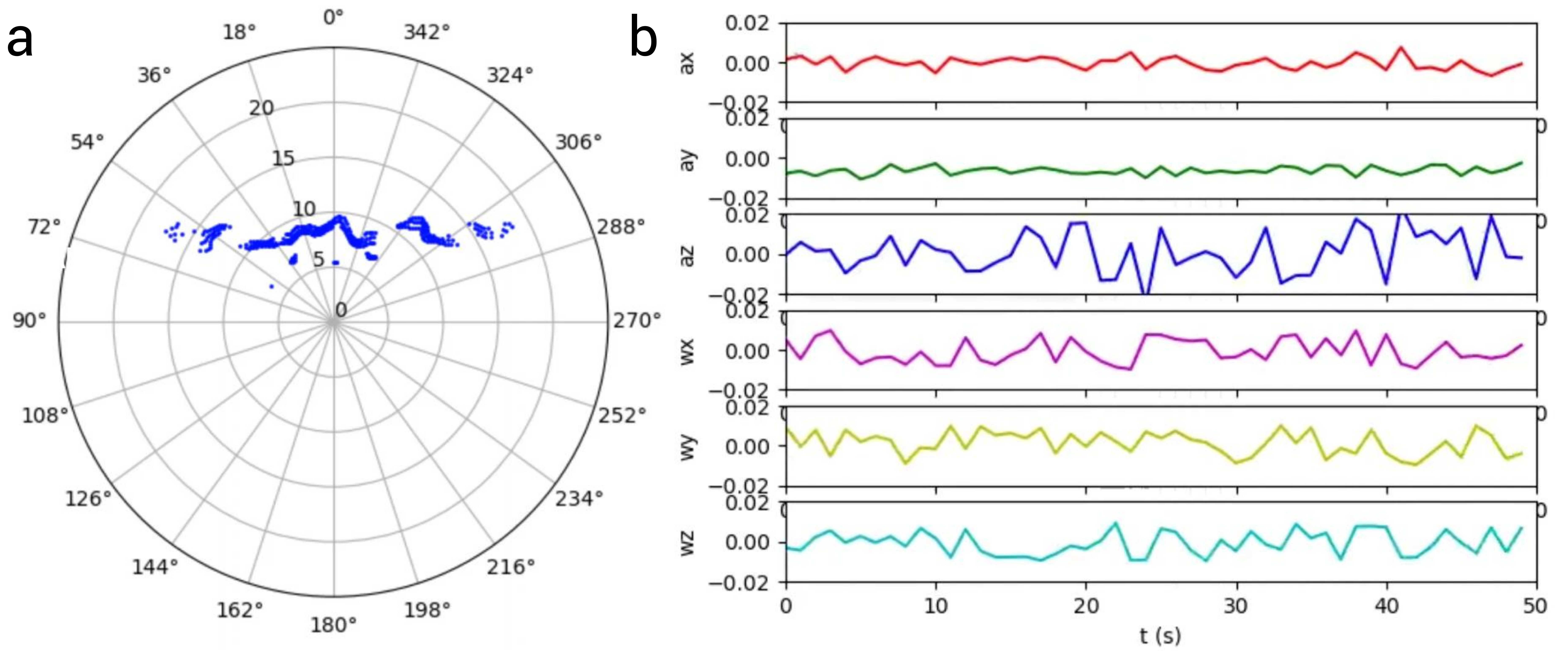}
		\caption{Sensor Sample Plot (a)Sonar Sample Plot (b) IMU Sample Plot from top: $a_x$, $a_y$, $a_z$ are linear acceleration in m/$s^2$, $w_x$, $w_y$, $w_z$ are angular velocity in rad/s. X-axis is time in seconds}
		\label{fig:sensor_sample_plot}
	\end{figure}
 
\subsection{Sensor deployment}
For underwater tasks, two of the most commonly used sensors are the IMU (Inertial Measurement Unit) and the SONAR (Sound Navigation and Ranging). Since we used a game engine, we have all the ground truth for the position data for all objects. 
\subsubsection{Modeling IMU}
IMU measurements are generated using the ground truth position of the rover from our simulation environment, Blender. Further noise augmentation is done in post-processing which incorporates various noise models. \newline 
\textbf{Measurement noise :} 
The IMU error model takes into account the bias, angle and velocity random walks, and bias instability correlation. Depending on the accuracy of the sensor used, we have defined three error models, ranging from low to high noise. 
\begin{equation}
    IMU_{sim} = IMU_{GT} + IMU_{bias} + IMU_{noise} 
\end{equation}
\begin{equation}
    Acc_{noise} = Acc_{(velocity\_random\_walk)} + Acc_{(vibration)}
\end{equation}
\begin{equation}
    Gyro_{noise} = Gyro_{(angle\_random\_walk)} + Gyro_{(vibration)}
\end{equation}
where the vibration model can be defined as a white Gaussian, sinusoidal, or power spectral density.

\subsubsection{SONAR}
Sound navigation and ranging (SONAR) sensor is widely used for underwater state estimation and mapping. As such, it should be an integral part of any underwater simulation environment. We use BLAINDER \cite{reitmann2021blainder} package to simulate the SONAR sensor model. The BLAINDER is an add-on package for a 3D modeling program Blender, which enables the creation of semantically annotated point cloud data in virtual 3D settings and streamlines the development of training data for a variety of areas. It concentrates on the traditional depth-sensing methods of Light Detection and Ranging (LiDAR) and SONAR.
\begin{figure}[b!]
		\centering      
		\includegraphics[width=\linewidth]{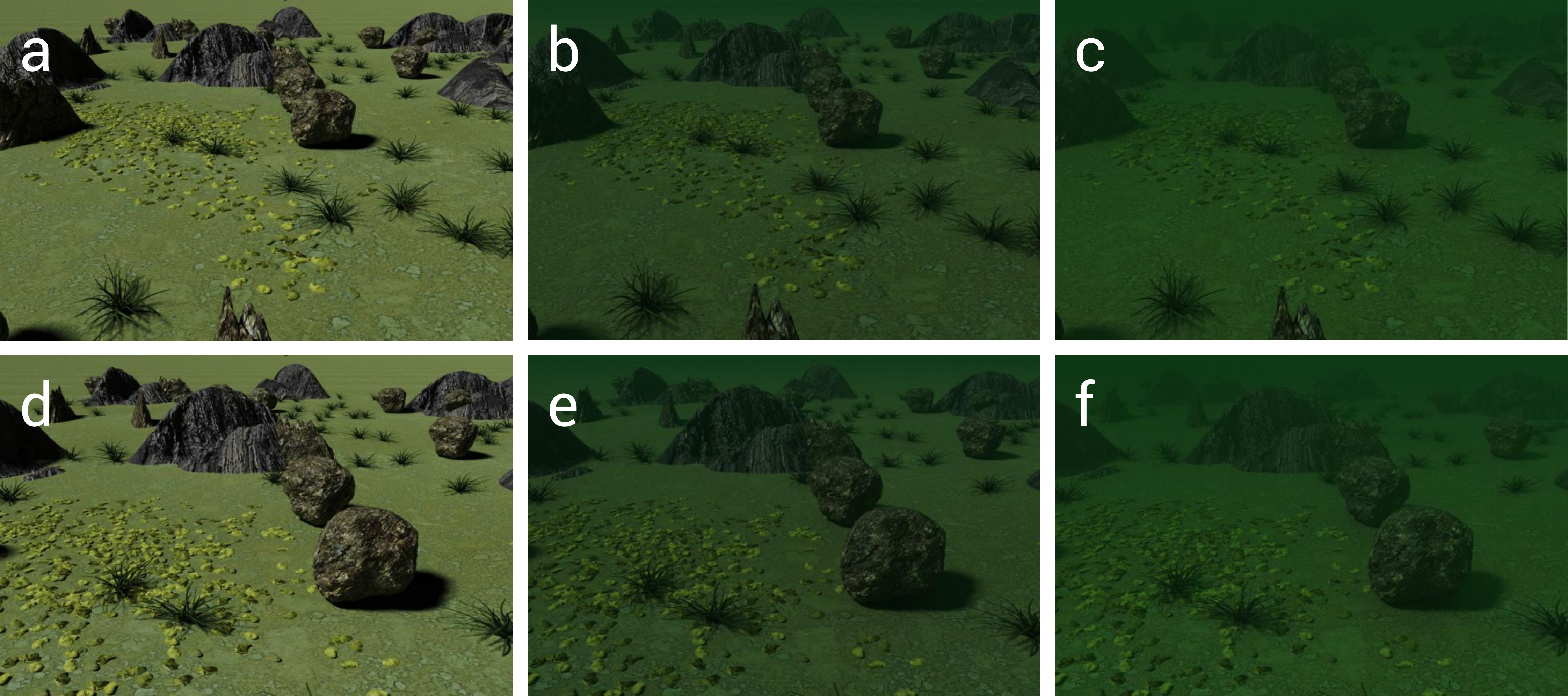}
		\caption{Water effect  Each row left to right: Simulation image with no water volume, simulation image with water volume, and simulation image with water volume that has more particles }
		\label{fig:water_effect}
	\end{figure}
The most deterministic term for the simulation of the sonar measurement is the velocity of propagation. The velocity is not homogeneous when traveling within a body of water.  The depth $D [m]$, temperature $T [^\circ C]$, and the salinity $S [\%]$, are all important factors that determine the velocity $c[m/s]$. An approximation formula for the calculation of $c$ is defined as follows \cite{IntroSonar}:
\begin{equation}
\begin{split}
{c} &=1449.2+4.6T - 0.055T^2 + 0.00029T^3 \\
&+ (1.34 - 0.010T)(S - 35) + 0.016D
\label{eq:c}
\end{split}
\end{equation} 
More details of the Sonar can be found in BLAINDER. \cite{reitmann2021blainder}
\subsection{Water volume}

A crucial aspect of the underwater simulation environment is the scattering properties of the water volume, as this is the main distinction between an above-water and underwater scene.
In Blender environment, the render engine "CYCLE" uses path-tracing to generate images. In short, it passes light and receives light to generate images of different shades, brightness, colors, etc. By adding particles to the water volume, we get a photo-realistic underwater image. The color of the volume can also be changed to mimic the underwater environment's color scheme. 
As we can see in Figure~\ref{fig:water_effect}(a) and Figure~\ref{fig:water_effect}(b), it just looks like an above-water image when there are no particles added and no color set to the water volume. With added particles and color changes, Figure~\ref{fig:water_effect}(b) and Figure~\ref{fig:water_effect}(e) looks much more like an underwater image. 
The density of the particles in the water can be adjusted as illustrated in Figure~\ref{fig:water_effect}(c) and Figure~\ref{fig:water_effect}(f). It is definitely harder to see/recognize objects with water volume that has more particles. 


\begin{figure}[h!]
		\centering      
		\includegraphics[width=\linewidth]{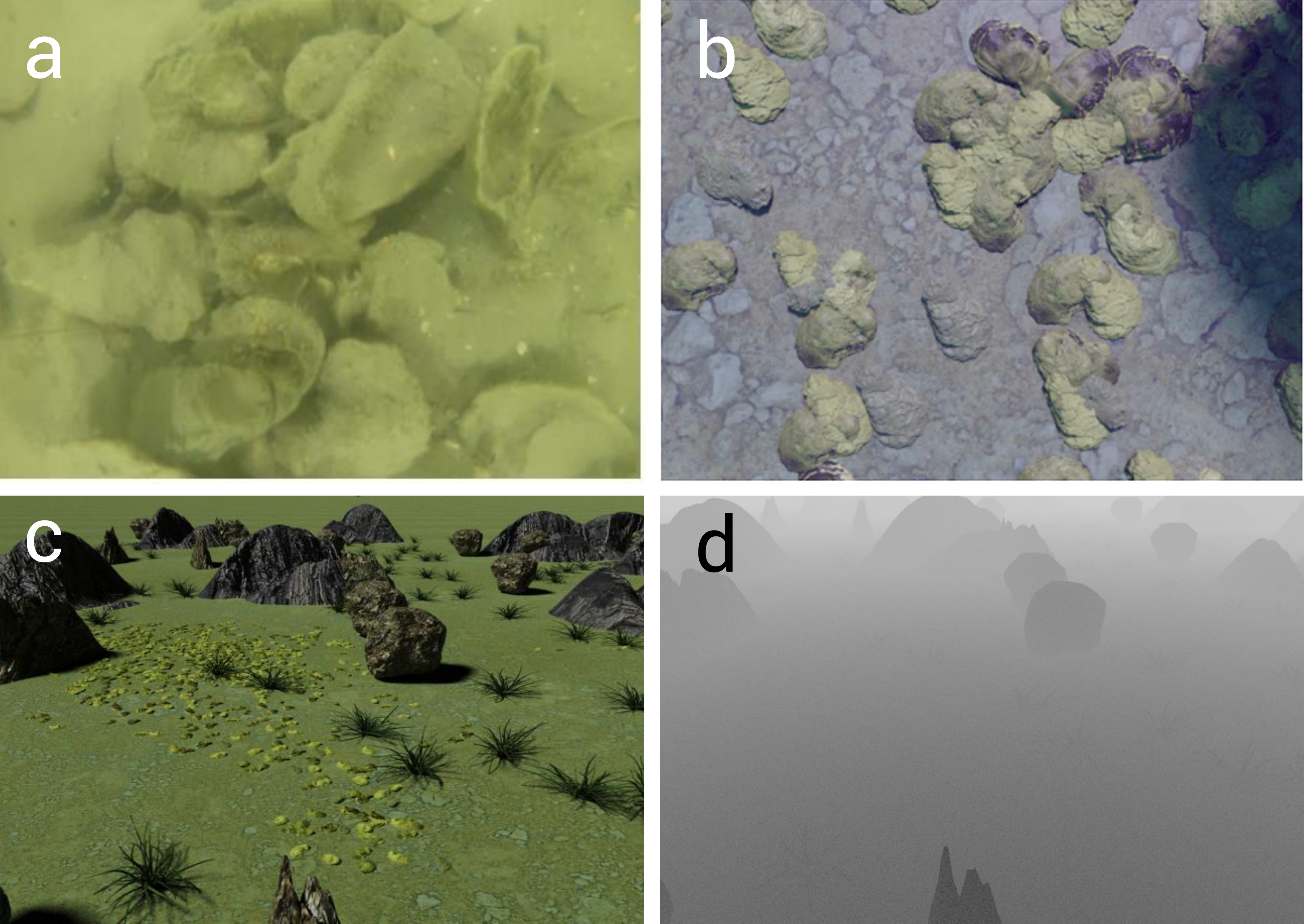}
		\caption{Underwater Images. (a) Real underwater images for the oysters. (b) Image from the down-facing camera on the ROV. (c) Image from the front-facing camera on the ROV. (d) Depth image from the front-facing camera on the ROV.}
		\label{fig:underwaterimages}
	\end{figure}
\subsection{Details of Simulation}
Real textures similar to Figure \ref{fig:underwaterimages}(a) of the seabed and oysters are used to create the simulation. In addition, we include some real 3D models of the stones and the rocks of various sizes to add more non-oyster objects to the scene. 
\begin{figure*}[h!]
		\centering      
		\includegraphics[width=0.9\linewidth]{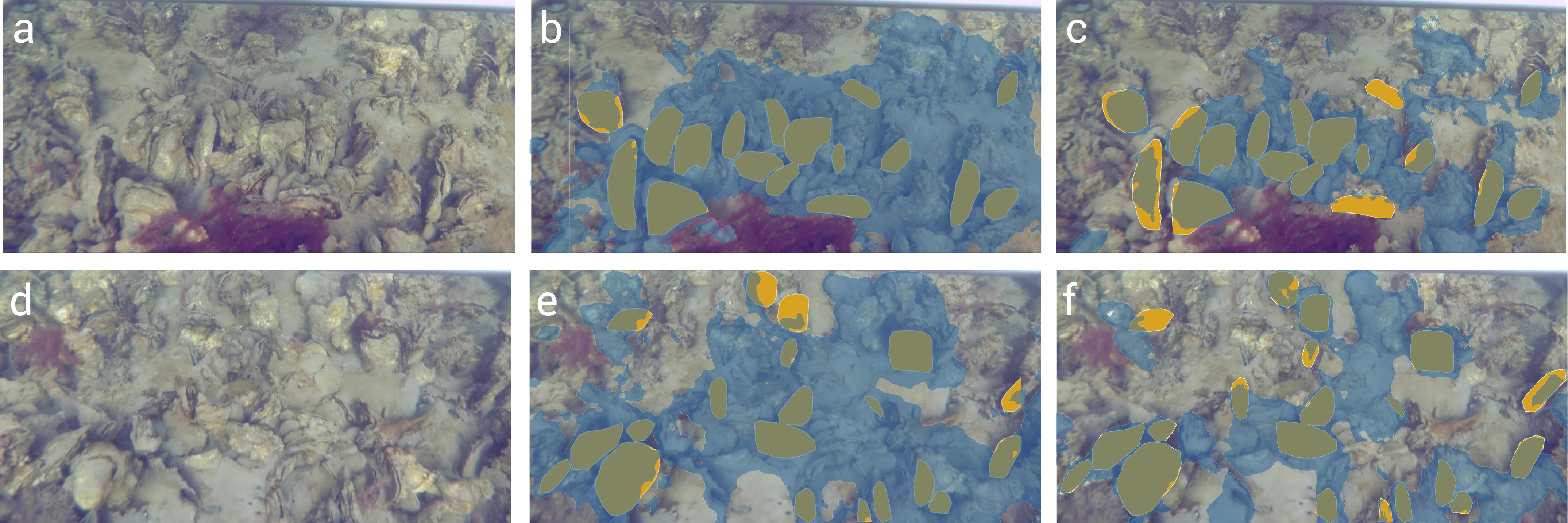}
		\caption{Oyster Detection Result. Each row left to right: Input image, output of the network when
trained using only real data, output of the network (which we call OysterNet)
when trained using real data augmented with our synthetic data. Yellow
represents the oyster segmentation ground truth and the blue is the predicted
segmentation result. Notice how the number of false positives and false
negatives drop significantly when the training data is augmented with our
synthetic data, All the images in this paper are best seen in color on a
computer screen at 200\% zoom.}
\label{fig:oyster_detection}
\end{figure*}
Water volume is included and we set the color scheme to green to simulate the environment for some oyster reefs. The color scheme, illumination level, and turbidity level of the water volume can also be adjusted based on needs. Four different cameras are utilized, which include third person view (Figure~\ref{fig:p1}), down-facing camera on the ROV (Figure~\ref{fig:underwaterimages}(b)), front-facing camera on the ROV (Figure~\ref{fig:underwaterimages}(c)) and front-facing depth camera on the ROV (Figure~\ref{fig:underwaterimages}(d)). 

\section{Applications}
\label{section:app}
\textit{OysterSim} can be used in various aspects of the oyster reef monitoring application. The generated simulated data can be used for oyster detection. The simulated environment can also be used to study the algorithm for oyster reef exploration. Moreover, the ground truth for the locations and the visual data we generated can also be used to study visual simultaneous localization and mapping(V-SLAM) for oyster-reef locations. 

\subsection{Oyster Detection}
In order to monitor oyster reefs, oyster detection would be the first step. However, training a deep neural network would require a massive labeled dataset. Obtaining such a dataset  would be expensive and labor-intensive in underwater environments.
In \textit{OysterNet}[], we utilized \textit{OysterSim} to automatically generate simulated oyster data with its ground truth. Then we used a CUT\cite{park2020contrastive} network to transform simulated oyster data into synthetic oyster data. Without the synthetic oyster data, we are able to achieve an oyster detection IOU (Intersection over Union) Score of $18.88\%$. As we can see in Figure~\ref{fig:oyster_detection}(a), there is a lot of misprediction. Results without synthetic oyster data are shown in light blue color whereas ground truth in dark blue.

With the combination of the synthetic oyster data and real oyster data, we are able to train \textit{OysterNet} with 24.4\% IOU Score which is 35.1\% above the state-of-the-art. As we can see in Figure \ref{fig:oyster_detection}(b), the result we obtained is much better. The detection result from \textit{OysterNet} with synthetic oyster data is shown in orange.
More details of training and oyster modeling can be found in the paper\cite{lin2022oysternet}.
\begin{figure}[!h]
		\centering      
		\includegraphics[width=0.9\linewidth]{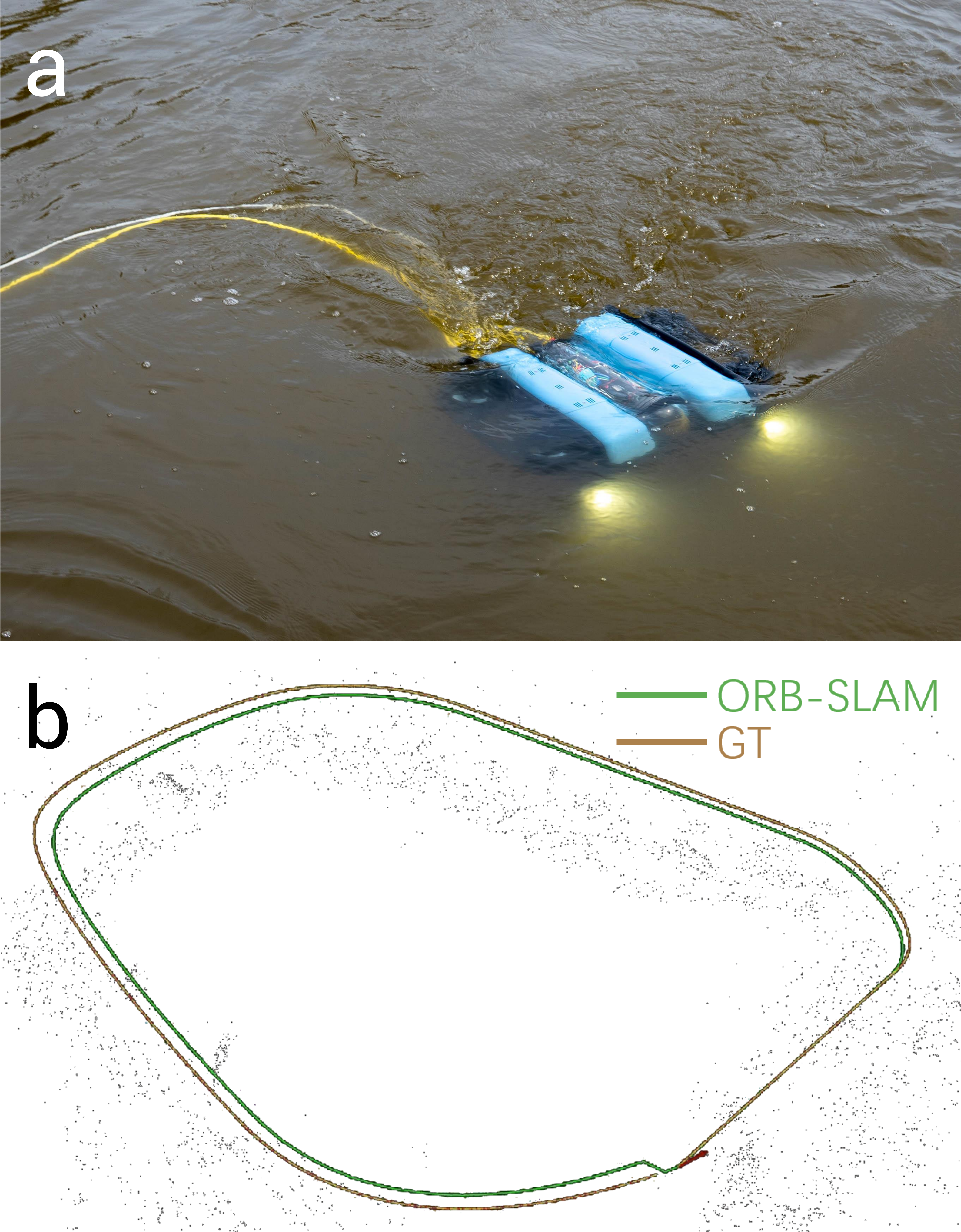}
		\caption{(a) The BlueROV collecting data of shallow oyster-reef. (b)  Trajectory generated from V-SLAM and ground truth. The green Trajectory is generated from running V-SLAM on the synthetic data. The Yellow Trajectory is the ground truth.}
		\label{fig:trajctory}
	\end{figure}
\subsection{V-SLAM}
To autonomously monitor the oyster reef with an underwater vehicle, a reliable localization and mapping capability is of vital importance. Particularly in the underwater environment, it is important to use the fusion of different sensor inputs - sonar and camera. As a proof-of-concept, we feed the simulated data that we generated from the simulation to one of the Visual Simultaneous localization and mapping algorithms, the ORB-SLAM~\cite{mur2017orb}, to show the localization result in Figure~\ref{fig:trajctory}. We can use the data we generated to compare different underwater SLAM algorithms with the generated ground truth. A lot of the SLAM algorithms are tested with estimated ground truth. Now they can be compared with the same dataset with different sensor data. There are numerous amount of sensors for underwater applications. Anyone who wants to utilize the open source code could add more sensors as they wish to. We would further add Doppler Velocity Log (DVL) to the list of sensors in the presented simulation. 
We plan on improving our mathematical model for the oysters and adding more textures to simulate the appearance of the oysters. In the current state, our simulation opens up opportunities for oyster reef researchers to develop and test navigation, and path planning algorithms with oyster-based localization.

\begin{figure}[!h]
		\centering      
		\includegraphics[width=0.95\linewidth]{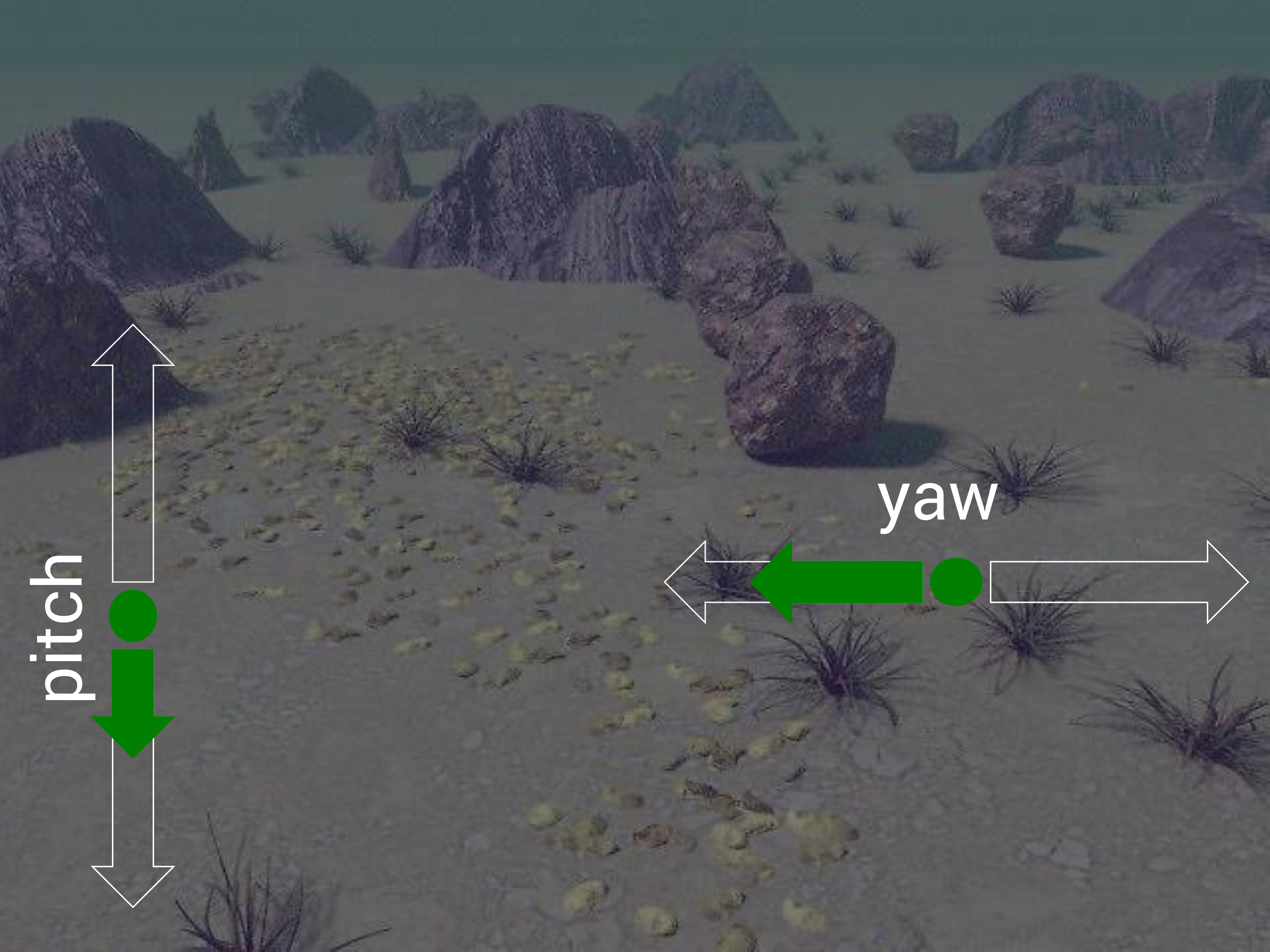}
		\caption{Exploration labeling. Pitch and yaw are labeled with 7 classes each}
		\label{fig:exploration}
	\end{figure}
\subsection{Oyster Reef Exploration}
To maximize the oyster information gained within a limited amount of time for a remotely operated vehicle (ROV), a novel exploration algorithm would be required. 

Inspired by vision-based navigation methods from Manderson et al.\cite{manderson2020vision} and Karapetyan et al.\cite{karapetyan2021human}, we are developing an algorithm that can mimic human controls for underwater oyster reef monitoring. Instead of collecting the dataset from the real oyster reef, \textit{OysterSim} can be utilized to generate synthetic data that can be used for human control labeling. 
As we can see from Figure~\ref{fig:exploration}, we divided the control of the robot into 14 discrete control classes with 7 labels for the pitch direction and 7 labels for the yaw direction. We have labeled 5000 underwater images generated from the simulation for learning a deep neural network to explore the oyster reef through behavioral cloning. The result is currently under development. 

\section{Conclusion}
\label{section:conclusion}
This paper presents simulation environment for oyster reef monitoring - \textit{OysterSim}. It enables the development of many aspects of the monitoring process including but not limited to navigation, detection and mapping. We modeled the oyster using the mathematical model of the contour shapes and deployed multiple sensors in \textit{OysterSim}. Water volume is also introduced for photo-realistic images. We applied our simulation in three different applications: oyster detection, oyster reef exploration, and visual simultaneous localization and mapping. \textit{OysterSim}can contribute to more aspects of oyster reef monitoring and opens up opportunities for more efficient operations. Oyster density map can be constructed based on the \textit{OysterSim} which can later be used to develop novel algorithm for path planning of oyster harvesting.

\section*{ACKNOWLEDGMENT}
This work is supported by USDA NIFA Sustainable
Agricultural Systems (SAS) Program (Award Number:
20206801231805).
\bibliographystyle{IEEEtran}
\bibliography{refs}
\nocite{*}

\end{document}